\documentclass[runningheads]{llncs}
\usepackage{graphicx}
\usepackage{amsmath,epsfig}
\usepackage[misc]{ifsym} 
\usepackage{multirow}
\usepackage{amssymb} 
\usepackage{colortbl}
\usepackage{xcolor}
\usepackage{tikz}
\usepackage{pifont}
\usepackage{array}
\usepackage{booktabs}
\usepackage{caption}
\usepackage{subcaption}
\usepackage{tikz}

\DeclareRobustCommand{\colorcircle}[1]{\tikz[baseline]{\draw[fill={#1}, draw=black] (0,0) circle (0.1cm);}}

\begin{document}
\title{CoSTL: Comprehensive Spatial-Temporal Representation Learning for Moment Retrieval and Highlight Detection
\vspace{-10pt}
\thanks{These authors contributed equally. $^{\dagger}$ Corresponding author. }
}

\titlerunning{CoSTL: Comprehensive S-T Representation Learning for MR and HD}

% \author{Xin Dong$^{1,2}$, Wenjia geng$^{1}$, Wenfeng Deng$^{2,\dagger}$, Yansong Tang$^{1,\dagger}$   \\
% $^1$Shenzhen International Graduate School, Tsinghua University $^2$Pengcheng Laboratory \\
%     dong-x23@mails.tsinghua.edu.cn
% }

\author{Xin Dong$^{*}$\inst{1,2} \and
Wenjia Geng$^{*}$\inst{1} \and
Wenfeng Deng\inst{2} \and
Yansong Tang$^{\dagger}$\inst{1}
\institute{Shenzhen International Graduate School, Tsinghua University \and
Pengcheng Laboratory}
}

\authorrunning{X. Dong et al.}

\maketitle              % typeset the header of the contribution

\begin{abstract}
Video Moment Retrieval (MR) and Highlight Detection (HD) are  crucial tasks in video analysis that aim to localize specific moments and estimate clip-wise relevance based on a given text query. Recent approaches treat them as similar video grounding tasks and use the same architecture to solve them. These tasks require both fine-grained comprehension at the image level and high-level temporal understanding across the entire video. Exsiting approaches have primarily focused on temporal modeling using frame-level features, often neglecting the rich visual information related to the text query within individual frames. This oversight leads to inaccurate grounding results. To address this limitation, we propose a \textbf{Co}mprehensive \textbf{S}patial-\textbf{T}emporal Representation \textbf{L}earning Framework (CoSTL), which captures both fine-grained image-level information and temporal dynamics. Specifically, CoSTL incorporates a text-driven progressive fine-grained image encoder, performing a two-step text-driven knowledge extraction process to learn fine-grained spatial representations. Furthermore, a multi-scale temporal perception module captures comprehensive spatial-temporal representations, enhancing the model's ability to process temporal dynamics. We demonstrate state-of-the-art performance on four public benchmarks: QVHighlights, Charades-STA, TACoS, and TVSum.


\end{abstract}    
\section{Introduction}
\label{sec:intro}
Video understanding has become increasingly popular and important in recent years. With the rapid growth in the number of available videos, accurately extracting key information from them has become a challenging problem. Consequently, moment retrieval \cite{2dtan,moment1,tall,ego4d} and highlight detection \cite{highlightgif,highlightjoint,highlightless,highlightyao2016} have emerged as significant topics in the field of video analysis, playing crucial roles in intelligent retrieval, automated video editing, and a variety of other applications~\cite{Wang2024LocalizationAwareMR,yang2022lavt}. Given the shared characteristics of these two tasks, a unified framework \cite{univtg,umt,mdetr,cgdetr} is often employed to address them together.

\begin{figure}[!t]
\centering
%\vspace{-10pt}
\includegraphics[width=0.9\linewidth,height=0.6\linewidth]{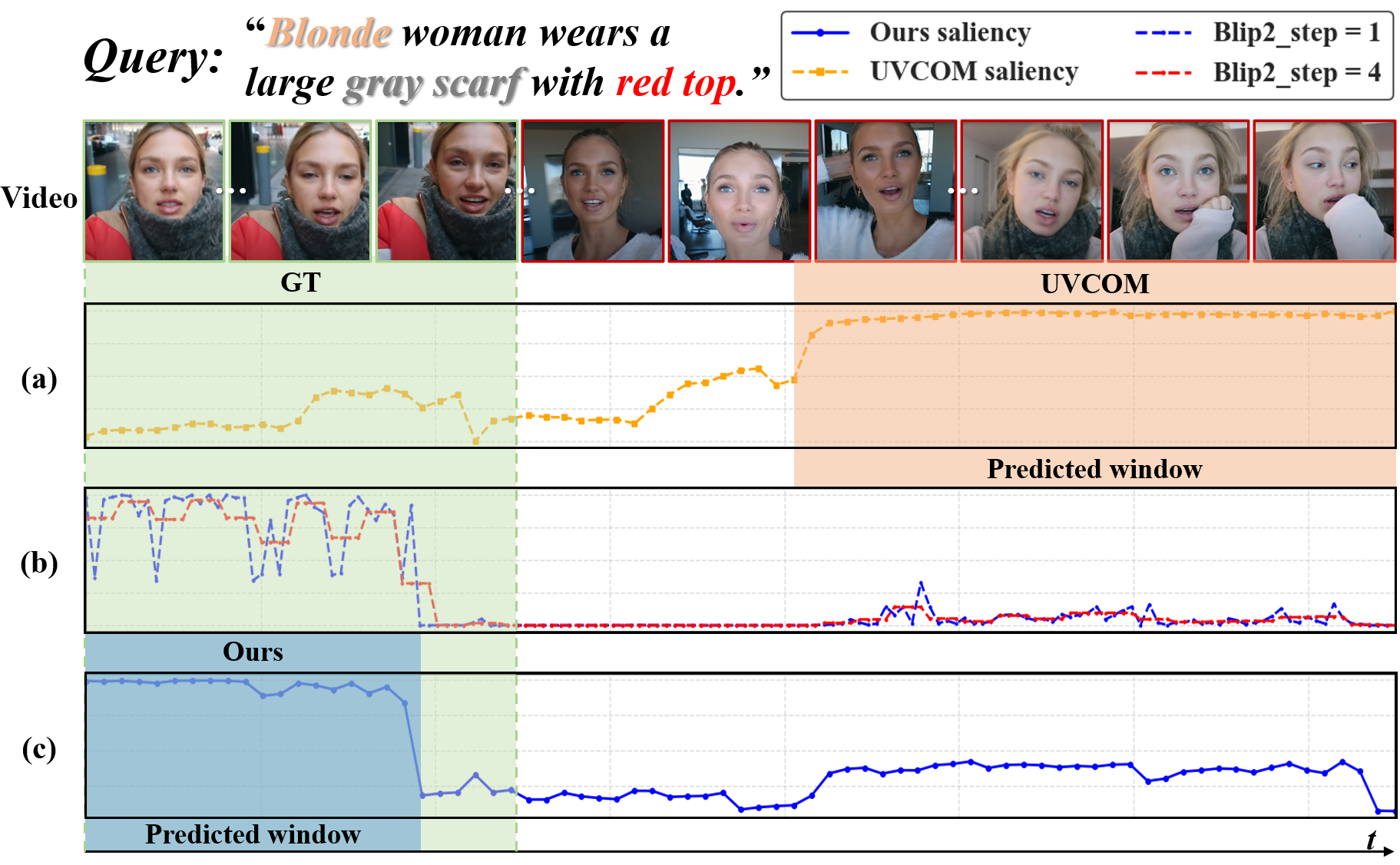}
    \caption{Comparison of highlight saliency scores across different methods: (a) denotes the previous state-of-the-art UVCOM; (b) illustrates the image-text matching scores of MLLM at various time scales and (c) presents our method. The results in (a) indicate that the previous approach struggles to capture fine-grained spatial details, while (b) highlights the temporal dynamics.}
    \label{fig:teaser}
    \vspace{-16pt}
\end{figure}
% \vspace{-10pt}
Given a text query, moment retrieval aims to localize the corresponding accurate moment within the entire video, and highlight detection also has to provide a saliency score for each related video frame. These two tasks require a comprehensive understanding of multi-granularity information. As illustrated in Fig. \ref{fig:teaser}, two main problems exist in these tasks: 
\textbf{(i) Fine-grained Visual Understanding at the Spatial Level}: Accurate localization requires precise comprehension of visual details within each frame. For example, to locate the moment corresponding to ``Blonde woman wears a large gray scarf with a red top", the model must identify fine-grained elements like the ``red top". The previous state-of-the-art method UVCOM \cite{uvcom}, as shown in Fig. \ref{fig:teaser} (a),  produces entirely incorrect results due to insufficient spatial detail extraction, which only adopts frame-level features. 
    \textbf{(ii) Temporal Dynamic Modeling at the Temporal Level}: While fine-grained spatial understanding is crucial, it is insufficient on its own. Fig. \ref{fig:teaser} (b) demonstrates this by employing a fine-grained multimodal language model (BLIP2) \cite{blip2} to calculate per-frame saliency scores. Despite capturing detailed information, this approach struggles to model the overall action due to its solely spatial focus. The resulting per-frame saliency scores (blue line) exhibit significant and unrealistic jumps between frames. However, these fluctuations are mitigated when considering a larger temporal context (red line), highlighting the importance of temporal modeling.
% \textbf{(i) fine-grained visual information understanding in spatial level}: to localize the sentence ``Blonde woman wears a large gray scarf with a red top.", the model first must have a fine-grained understanding of each frame, such as identifying the ``red top". The previous STOA, as shown in Figure 1(a), produces totally wrong result due to the lack of spatial detail extraction. \textbf{(ii) temporal dynamic modeling within the video}: the sorely spatial information extraction is not enough, as shown in Figure 1(b), we utilize a fine-grained multi-modal language model BLIP2 to calculate saliency score per frame. Though detailed information is modeled, the modeling of the entire action cannot be modeled from the spatial dimension alone. The result of a single frame (blue line) has a large jump between frames, which is unreasonable. However, when we calculate on a larger time dimension, such jump is alleviated.

% \vspace{-15pt}
To address this issue, we propose a model that captures optimal spatial-temporal representations corresponding to the text query, spanning from the image level to the video level.
At the image level, our model aims to extract fine-grained spatial features. Inspired by recent advancements in multimodal large language models, we introduce a text-driven progressive fine-grained frame encoder. This encoder employs a two-step text-driven feature extraction process: the first step involves extracting the desired features from all spatial patches, while the second step further interacts with the text to obtain the representation of the entire video frame.
After obtaining fine-frained spatial-level representations, then we address the aforementioned temporal dynamics challenge with a multi-scale temporal perception module at the video level. This module captures action information at both coarse and fine granularities, modeling both macro-level actions and precise temporal details. This multi-scale approach facilitates the learning of richer and more robust spatial-temporal representations. Fig. \ref{fig:teaser} (c)  shows that our method can get correct prediction results for complex scenes.

%To address this issue, we develop a model capable of capturing optimal spatial-temporal corresponding to the text query, ranging from the image spatial level to the video temporal level. At the image spatial level, the model needs to extract the fine-grained features most relevant to the text. The recent emergence of multimodal large models has provided numerous ideas for modeling fine-grained image information. Inspired by the classic multimodal large model BLIP2, we propose a text-driven progressive fine-grained frame encoder, which extracts the multimodal features of video frames most relevant to the text through a two-step text-driven feature extraction module. The first step is to extract the desired dimensions from all spatial patches, and the second step is to further interact with the text to obtain the representation of the entire video frame. At the video temporal level, to solve the temporal dynamic issue in last paragraph, we designed a multi-scale temporal perception module, which models macro action information at a coarse granularity and captures precise action information at a fine granularity. Modeling multi-scale temporal information enables the model to learn better spatial-temporal representations.

% \vspace{-10pt}
We conduct extensive experiments on four popular Moment Retrieval and Highlight Detection benchmarks to validate the effectiveness of our model. The results demonstrate that our model outperforms existing methods on all benchmarks. Overall, our contributions are summarized as follows:
\vspace{-5pt}
\begin{itemize}
    \item We find the primary bottleneck affecting the performance of MR/HD is the neglect of fine-grained visual information extraction and temporal dynamics modeling. Based on this observation, we designed CoSTL, a comprehensive spatial-temporal representation learning framework that effectively learns optimal image-spatial and video-temporal level knowledge.
    \item In CoSTL, a text-driven progressive fine-grained image encoder is proposed to learn fine-grained spatial-level embedding, followed by a multi-scale temporal perception module to model temporal dynamics within the video.
    \item Extensive experiments across moment retrieval and highlight detection on four benchmarks demonstrate the significance of our method.
\end{itemize}

\section{Related Works}
\label{sec:related}
\textbf{Video grounding (VG)} refers to extracting specific moments and relevant clips from videos based on a given text query~\cite{Moon2026CVACV,Yang2025TimeexpertAE}. With the increasing volume of video content, which is central to applications ranging from online streaming to automated video editing, there is a growing need for efficient VG.Two fundamental tasks in VG are \textbf{Moment Retrieval (MR)} \cite{zhang2021multi,yuan2019semantic,wang2021structured,univtg} and \textbf{Highlight Detection (HD)} \cite{gygli2016video2gif,jiao2018video,badamdorj2021joint,liu2022umt,univtg}. Specifically, MR focuses on localizing a specific moment in the video that aligns with a given text query while HD aims to identify and score multiple important segments that capture significant actions or events within the video. Recent works discovers new frames works to solve MR and HD simultaneously since they share many similar properties \cite{univtg,liu2022umt,uvcom}. 

\textbf{Moment Retrieval. }
Previous methods can be categorized into two types \cite{uvcom}: proposal-based and proposal-free. Proposal-based methods typically follow a two-stage matching process \cite{zhang2021multi,yuan2019semantic,wang2021structured}. In the first stage, candidate proposals are generated using multi-scale sliding windows \cite{liu2018attentive,hendricks2018localizing}. In the second stage, these proposals are paired and ranked by explicitly modeling channel-level or sequence-level interactions. Despite their promising performance, these methods often suffer from high computational overhead due to the need to generate a large number of proposals, which can negatively affect efficiency and complicate the subsequent matching process. In contrast, proposal-free methods \cite{li2021proposal,mun2020local,rodriguez2020proposal,tang2021frame,yuan2019find,zhang2020span} bypass the proposal generation step entirely. Instead, they directly regress the start and end boundaries of the target without requiring predefined proposals. These methods typically decompose both video and language into structured hierarchies, such as word-phrase-sentence or event-actions-objects. This hierarchical representation allows for systematic encoding of the semantic content in both video and query text. The semantic correlation between text and vision is then learned through interactions among these multi-modal semantic units.

\textbf{Highlight Detection. }
Early HD videos typically feature a clear theme indicated by their titles or topics \cite{youtbe,tvsum}. Recently, QVHighlight \cite{mdetr} introduced a benchmark for joint MR and HD, allowing users to generate various highlights from a single video based on different text queries. Many prior works \cite{highlightgif,highlightless,highlightyao2016,yu2018deep,youtbe} adopt a ranking-based approach, where video segments are ranked by importance, with higher scores for more relevant segments. In contrast, Xiong et al. \cite{highlightless} focus on using only video-level tags as weak supervision. Recently, Liu et al. \cite{r2_tuning} proposed a parameter-efficient framework for video temporal grounding using multi-layer CLIP features, achieving top results.

\section{Method}
\subsection{Overview}
\begin{figure*}
\centering
%\vspace{-10pt}
\includegraphics[width=1\linewidth]{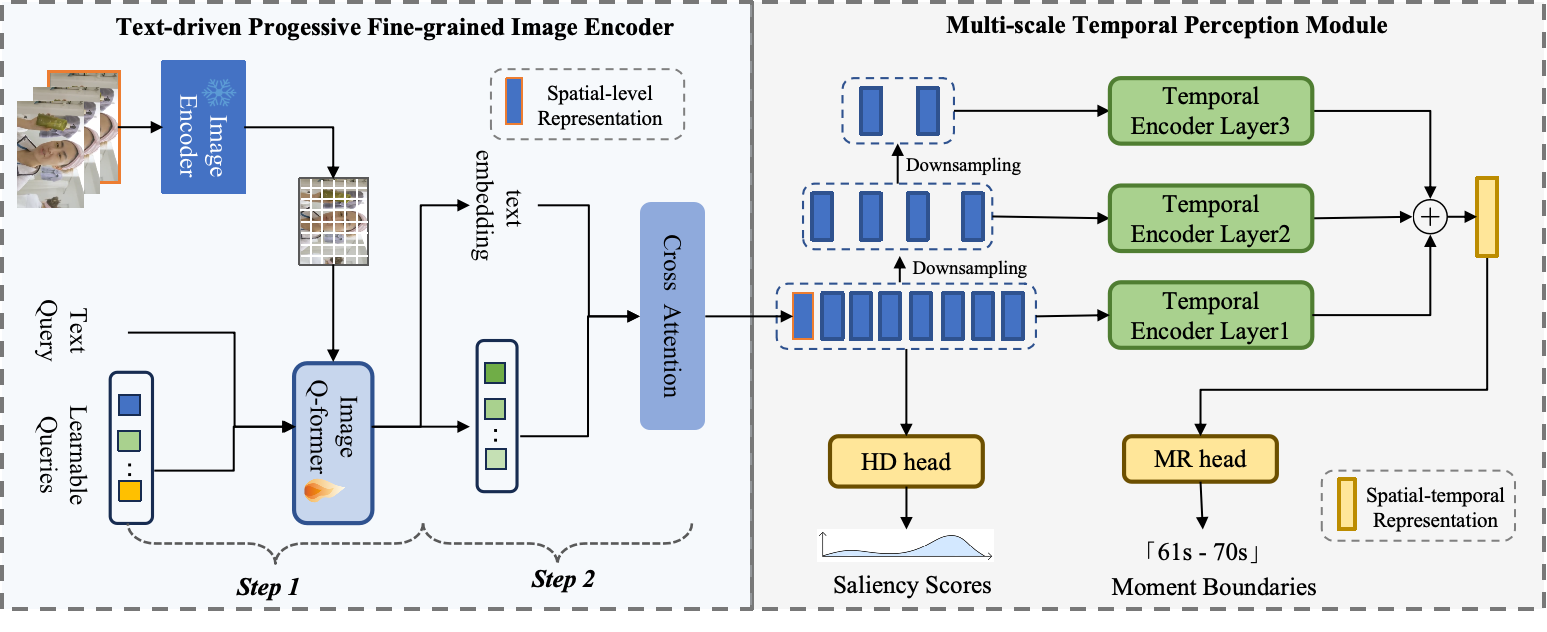}
    \captionof{figure}{Overall architecture of our framework. The input video and query are first encoded by a text-driven progressive fine-grained image encoder, obtaining precise spatial-level representations for each frame. These representations are then passed through a multi-scale temporal perception module to capture temporal dynamics, followed by different prediction heads to get final results.}
    \label{fig:pipeline}
    \vspace{-13pt}
\end{figure*}

Given a video $V=\left \{ v_{i}  \right \} _{i=1}^{H} $ and a natural language query $T=\left \{ t_{i}  \right \} _{i=1}^{L} $, where $H$ and $L$ are the number of video frames and text tokens. The goal of moment retrieval is to give multiple relevant moment offsets, presented in the form of center coordinate and duration $d_{i} = [d_{i}^{s}, d_{i}^{e}] \in R^{2} $, it also requires foreground indicator $f_{i} \in \left \{ 0,1 \right \} $, indicating the corresponding $d_{i}$ is valid or not. Highlight detection aims to generate the saliency score $s_{i} \in [0,1]$ for each frame in the video, which represents the relevance score between the visual content of clip $v_{i}$ and text query $T$.
Fig. \ref{fig:pipeline} provides an overview of our proposed method for moment retrieval and highlight detection. It offers a unified framework that can both conduct fine-grained spatial information extraction and multi-scale temporal perception, producing optimal spatial-temporal representations for these two tasks. We achieve this through two key modules: a Text-driven Progressive Fine-grained Image Encoder (TDE) operating at the spatial level and a Multi-scale Temporal Perception Module (MTP) focusing on the temporal dimension. The TDE extracts detailed spatial features, while the MTP captures dynamic temporal patterns at multiple scales. The following sections detail the design and functionality of both the TDE and MTP modules.

% \vspace{10pt}

\subsection{Text-Driven Progressive Fine-grained Encoder}
The Text-driven Progressive Fine-grained Encoder (TDE) extracts text-relevant fine-grained image features from a given raw image through a two-step interaction with the provided text. Compared to directly employing global features from a Vision Transformer (ViT) \cite{vit}, TDE preserves more fine-grained information pertinent to the text. Given a raw video as input, we first extract initial visual features, represented as $F_{v} \in R^{B\times N\times T\times D_{v} } $, encompassing all spatial information. Here, $B$ denotes the batch size, $N$ the number of video frames, $T$ the number of patches per frame, and $D_{v}$ the dimension of the visual features. We then initialize learnable queries $F_{q} \in R^{B\times N\times M\times D_{q} } $, where $M$ represents the number of learnable queries and $D_{q}$ is their dimension. The first interaction with the text employs the same Q-Former architecture as in BLIP2 \cite{blip2}. This step aims to adaptively extract text-relevant spatial features from the patch-level visual representations $F_{v}$. Specifically, the learnable queries interact with each other through self-attention layers, then interact with the visual features through cross-attention layers. This process can be formulated as:
\begin{equation}
F_{v}^{'} =MLP(MHCA(F_{v},MHSA(F_{q})))
\end{equation}
where $MHCA$ represents multi-head cross-attention and $MHSA$ represents multi-head self-attention, $F_{v}^{'} \in R^{B\times N\times M\times D_{q}} $, which compress the spatial information dimension from $T$ to $M$. Besides, the learnable queries additionally interact with the text through the same self-attention layers.
\begin{equation}
F_{t} =MLP(MHSA(T))
\end{equation}
where $F_{t} \in R^{B\times N\times D_{t} }$, and $D_{t}$ denotes the dimension of text features. $F_{v}^{'}$ learns text-related representation through the same self-attention layers.

In the first step, we utilize an Image Q-Former to reduce the spatial dimension of the image features to $M$, which aims to retain fine-grained information relevant to the text while distilling the most important spatial features. Subsequently, we conduct second interaction with the text, aiming to obtain a global spatial-level representation that effectively captures the text's semantic content.
\begin{equation}
F_{v}^{''} = softmax\left ( \frac{\left ( w_{v}F_{v}^{'}  \right ) ^{\top } w_{t}F_{t}}{\sqrt{d_{k} } }  \right )\cdot F_{v}^{'}\in R^{B\times N\times D_{q}} 
\end{equation}
where $w_{v}$ and $w_{t}$ are learnable matrices for projecting features and $d_{k}$ is projected feature dimension. 

% \vspace{-5pt}
\subsection{Multi-Scale Temporal Perception Module}
% In the last section, we have modeled the text and visual information within the spatial dimension, yielding fine-grained spatial features most relevant to the text for each video frame. However, modeling based only on spatial representation will cause frame jump problems, so we further conduct temporal modeling. Specifically, we employ a multi-scale temporal perception module to capture comprehensive temporal dynamics, enabling the effective modeling of actions. 
Multi-scale temporal perception allows us to capture both short-term and long-term temporal dependencies, crucial for understanding complex actions and events within the video. Specifically, we operate over three distinct temporal scales: $S = \left \{ s_1, s_2, s_3 \right \}  $. For each scale $s_i \in S$, we first generate a corresponding video representation by temporally downsampling the original video features:
\begin{equation}
F_{v}^{s_{i}} =Maxpooling(F_{v}^{''})^{\left ( s_{i}  \right ) } \in R^{B\times (N/s_{i} )\times D_{q}}
\end{equation}
To capture temporal dependencies at each scale, the video features ${F_v^{s_1}, F_v^{s_2}, F_v^{s_3}}$ are individually processed by a standard Transformer encoder layer: 
\begin{equation}
\hat{F _{v}^{s_{i} } } = FFN(MHSA(F _{v}^{s_{i} } ))
\end{equation}
This yields scale-specific video representations $\hat{F _{v}^{s_{1} } }$, $\hat{F _{v}^{s_{2} } }$, $\hat{F _{v}^{s_{3} } }$. Then we apply linear temporal interpolation to each feature $F_v^{s_i}$ to ensure consistent temporal dimensions across all scales:
\begin{equation}
\tilde{F _{v}^{s_{i} }} = upsampling\left ( \hat{F _{v}^{s_{i} } } \right ) \in R^{B\times N\times D_{q}} 
\end{equation}
 These multi-scale representations are then fused to the final spatial-temporal representation $\tilde{F_{v}}$, capturing a comprehensive view of the temporal dynamics.
\begin{equation}
\tilde{F_{v}} = {\tilde{F _{v}^{s_{1} } }+\tilde{F _{v}^{s_{2} } }+\tilde{F _{v}^{s_{3} } }}
\end{equation}

% where $\tilde{F_{v}}$ denotes the fine-grained spatial-temporal representation for these two tasks.

% \vspace{-5pt}
\subsection{Prediction Heads and Loss Function}
After obtaining spatial level representation $F_{v}^{''}$ and spatial-temporal representation $\tilde{F_{v}}$, we adopt different prediction heads and optimization objectives for these two tasks.

\textbf{Boundary Heads for Moment Retrieval.} To predict moment boundaries from $\tilde{F_{v}}$, we employ two parallel branches of convolutional layers. Each branch consists of three 1x3 convolutional layers followed by a ReLU activation. The first branch has two output channels, outputing the predicted moment offsets $\tilde{d}_{i}$. The second branch has a single output channel followed by a Sigmoid activation function, producing a foreground indicator $\tilde{f}_i$. We supervise the foreground predictions $\tilde{f}_i$ using a binary cross-entropy loss.
\begin{equation}
\mathcal{L}_{\mathrm{f}}=-\lambda_{\mathrm{f}}\left(f_{i} \log \tilde{f}_{i}+\left(1-f_{i}\right) \log \left(1-\tilde{f}_{i}\right)\right)
\end{equation}
Then the combination of smooth L1 loss and weighted generalized IoU loss is used to supervise prediction moment offsets $\tilde{d}_{i}$:
\begin{equation}
\mathcal{L}_{\mathrm{b}}=\lambda_{\mathrm{L} 1} \mathcal{L}_{\text {SmoothL } 1}\left(\tilde{d}_{i}, d_{i}\right)+\lambda_{\text {iou }} w_{i} \mathcal{L}_{\text {iou }}\left(\tilde{d}_{i}, d_{i}\right)
\end{equation}
%where ${w}_{i}$ is the dynamic weight coefficient set for different iou intervals. When the iou is higher, the weight is larger, indicating the prediction of high iou requires higher accuracy.
where $\lambda_{L1}$ and $\lambda_{GIoU}$ are weighting factors for the respective loss components, and $w_i$ represents a dynamic weight applied to the GIoU loss. Higher GIoU values result in larger weights $w_i$, emphasizing the importance of precise predictions for moments with high overlap. %This weighting strategy encourages the model to prioritize accurate boundary predictions for moments already well-localized.

\textbf{Saliency Heads for Highlight Detection} 
As for highlight detection, given $F_{v}^{''}$, we apply linear head to get saliency score $\tilde{s}_i$. Then we also adopt binary cross-entropy loss $\mathcal{L}_{\mathrm{h}}$ to supervise prediction $\tilde{s}_{i}$.
\begin{equation}
\mathcal{L}_{\mathrm{h}}=-\lambda_{\mathrm{s}}\left(s_{i} \log \tilde{s}_{i}+\left(1-s_{i}\right) \log \left(1-\tilde{s}_{i}\right)\right)
\end{equation}
Generally, the total loss is as follows:
\begin{equation}
\mathcal{L}_{\text {total }}=\mathcal{L}_{f}+\mathcal{L}_{b}+\mathcal{L}_{\text {h}}
\end{equation}

% \vspace{-10pt}
% \subsection{Inference}
% We derive the prediction of moment retrieval based on $\tilde{d}_i$ and $\tilde{f}_i$, firstly we obtain the start and end timestamps $[d_{i}^{s}, d_{i}^{e}]$ from $\tilde{d}_i$, then we get the confidence score $\tilde{s}_i$ for this interval by accumulating all foreground scores $\tilde{f}_i$ for this time interval:
% \begin{equation}
% S_{i} =  {\textstyle \sum_{d_{i}^{s} }^{d_{i}^{e} }} \tilde{f}_i
% \end{equation}
% After that, we apply NMS with threshold $\theta _{IoU} = 0.7 $ to reduce duplicates. As for highlight detection, we directly use $\tilde{s}_i$ to calculate the metrics.

\section{Experiments}
% In this section, we conduct experiments on various benchmarks to evaluate our approach.
\subsection{Experiment Settings}
% Experiment Settings such as datasets, evaluation metrics and implementation details are introduced below.

\textbf{Datasets.}
We conduct experiments on four mainstream datasets on two tasks. The details of the datasets, tasks, labels, and domains are shown in Table \ref{tab:datasets}.
\begin{table}[t]
    \caption{Summary of four mainstream datasets. We show their tasks, labels, number of samples and domains.}
    \centering
    % \resizebox{0.85\columnwidth}{!}{  % 使表格宽度自适应当前栏宽度
        % \begin{tabular}{lccccc}
        \begin{tabular*}{\columnwidth}{@{\extracolsep{\fill}}ccccc}
            \toprule
            \textbf{Dataset} & \textbf{Task} & \textbf{Label} & \textbf{\# Samples} & \textbf{Domain} \\
            \midrule
            QVHighlights \cite{mdetr} & MR + HL & Interval + Curve & 10.3K & VLog, News \\
            Charades-STA \cite{tall} & MR & Interval & 16.1K & Indoor \\
            TACoS \cite{tacos} & MR & Interval & 18.2K & Kitchens \\
            TVSum \cite{tvsum} & HL & Curve & 50 & Web \\
            \bottomrule
        \end{tabular*}
    % }
    \label{tab:datasets}
    \vspace{-15pt}
\end{table}
\begin{table*}[h]
    \centering

    \caption{Video moment retrieval (MR) and highlight detection (HD) results on QVHighlights test split. 
    \colorcircle{rgb:blue,1;white,2} denotes CLIP\_B/32, 
    \colorcircle{rgb:green,1;white,2} denotes SlowFast R-50, \colorcircle{rgb:yellow,1;white,2} denotes audio features and
    \colorcircle{rgb:red,1;white,2} denotes BLIP2 (CLIP\_L) feature. The best and second-best metrics are marked with \textbf{bold} and \underline{underline}, respectively.}

    \begin{tabular}{cccccccccc}
        \toprule
        \multirow{3}{*}{\textbf{Method}} & \multirow{3}{*}{\textbf{Backbone}} & \multirow{3}{*}{\textbf{Pretrain}} & \multicolumn{5}{c}{\textbf{MR}} & \multicolumn{2}{c}{\textbf{HD}} \\ \cmidrule(lr){4-8} \cmidrule(lr){9-10}
        & & & \multicolumn{2}{c}{R1} & \multicolumn{3}{c}{mAP} & \multicolumn{2}{c}{$\geq$ Very Good} \\ \cmidrule(lr){4-5} \cmidrule(lr){6-8} \cmidrule(lr){9-10}
        & & & @0.5 & @0.7 & @0.5 & @0.75 & Avg. & mAP & HIT@1  \\ \midrule
        M-DETR\cite{mdetr} & \tikz\draw[fill={rgb:blue,1;white,2}, draw=black] (0,0) circle (0.1cm); \tikz\draw[fill={rgb:green,1;white,2}, draw=black] (0,0) circle (0.1cm); & \ding{55} & 52.89 & 33.02 & 54.82 & 29.40 & 30.73 & 35.69 & 20.88 \\
          UniVTG\cite{univtg} & \tikz\draw[fill={rgb:blue,1;white,2}, draw=black] (0,0) circle (0.1cm); \tikz\draw[fill={rgb:green,1;white,2}, draw=black] (0,0) circle (0.1cm); & \ding{55} & 58.86 & 40.86 & 57.60 & 35.59& 35.47 & 38.20 & 60.96 \\ 
          MH-DETR\cite{mhdetr} & \tikz\draw[fill={rgb:blue,1;white,2}, draw=black] (0,0) circle (0.1cm); \tikz\draw[fill={rgb:green,1;white,2}, draw=black] (0,0) circle (0.1cm); & \ding{55} & 60.05 & 42.48 & 60.75 & 38.18 & 38.38 & 38.22 & 60.51 \\ 
          EaTR\cite{eatr} & \tikz\draw[fill={rgb:blue,1;white,2}, draw=black] (0,0) circle (0.1cm); \tikz\draw[fill={rgb:green,1;white,2}, draw=black] (0,0) circle (0.1cm); & \ding{55} & 61.36 &45.79& 61.86& 41.91& 41.74& 37.15& 58.65  \\ 
        QD-DETR\cite{qddetr} & \tikz\draw[fill={rgb:blue,1;white,2}, draw=black] (0,0) circle (0.1cm); \tikz\draw[fill={rgb:green,1;white,2}, draw=black] (0,0) circle (0.1cm); & \ding{55} & 62.40 & 44.96 & 62.52 & 39.88 & 39.86 & 38.94 & 62.40 \\
        CG-DETR\cite{cgdetr} & \tikz\draw[fill={rgb:blue,1;white,2}, draw=black] (0,0) circle (0.1cm); \tikz\draw[fill={rgb:green,1;white,2}, draw=black] (0,0) circle (0.1cm); & \ding{55} & 65.43& 48.38& 64.51& 42.77& 42.86& 40.33& 66.21 \\
        UVCOM\cite{uvcom} & \tikz\draw[fill={rgb:blue,1;white,2}, draw=black] (0,0) circle (0.1cm); \tikz\draw[fill={rgb:green,1;white,2}, draw=black] (0,0) circle (0.1cm); & \ding{55} & 63.55 & 47.47 & 63.37 & 42.67 & 43.18 & 39.74 & 64.20 \\ 
        UVCOM\textsuperscript{*}\cite{uvcom} & \tikz\draw[fill={rgb:red,1;white,2}, draw=black] (0,0) circle (0.1cm); & \ding{55} & \underline{68.52} & \underline{51.94} & 65.40 & 44.06 & 43.78 & 40.64 & 65.42 \\ 
        R2-tuning\cite{uvcom} & \tikz\draw[fill={rgb:blue,1;white,2}, draw=black] (0,0) circle (0.1cm); & \ding{55} & 68.03 & 49.35 & \underline{69.04} & \underline{47.56} & \underline{46.17} & \underline{40.75} & 64.20 \\
        \midrule
        UMT\cite{umt} & \tikz\draw[fill={rgb:blue,1;white,2}, draw=black] (0,0) circle (0.1cm); \tikz\draw[fill={rgb:green,1;white,2}, draw=black] (0,0) circle (0.1cm); \tikz\draw[fill={rgb:yellow,1;white,2}, draw=black] (0,0) circle (0.1cm);& \checkmark & 60.83 & 43.26 & 57.33 & 39.12 & 38.08 & 39.12 & 62.39 \\ 
          UniVTG\cite{univtg} & \tikz\draw[fill={rgb:blue,1;white,2}, draw=black] (0,0) circle (0.1cm); \tikz\draw[fill={rgb:green,1;white,2}, draw=black] (0,0) circle (0.1cm); & \checkmark & 65.43 & 50.06 & 64.06 & 45.02& 43.63 & 40.54 & \underline{66.28} \\ 
        \midrule
        \textbf{CoSTL (Ours)} & \tikz\draw[fill={rgb:red,1;white,2}, draw=black] (0,0) circle (0.1cm); & \ding{55} & \textbf{75.43} & \textbf{54.24} & \textbf{71.48} & \textbf{48.45} & \textbf{46.68} & \textbf{41.36} & \textbf{71.59} \\
        \bottomrule
    \end{tabular}
    \label{tab:qvhighlight}
\end{table*}
% moment retrieval实验： charades and tacos数据集
\begin{table}[t]
    \centering
    \caption{Comparison of different methods on Charades-STA and TACoS datasets in terms of R@0.5 and R@0.7.}
    \small
    % \resizebox{0.56\columnwidth}{!}{
    % \begin{tabular}{ccccc}
    \begin{tabular*}{\columnwidth}{@{\extracolsep{\fill}}ccccc}
        \toprule
        \multirow{2}{*}{\textbf{Method}} & \multicolumn{2}{c}{\textbf{Charades-STA}} & \multicolumn{2}{c}{\textbf{TACoS}} \\ \cmidrule(lr){2-3} \cmidrule(lr){4-5}
        &R@0.5 & R@0.7 &R@0.5 & R@0.7  \\ \midrule
        2D TAN \cite{2dtan} & 46.02 & 27.50 & 27.99 & 12.82\\
        M-DETR \cite{mdetr} & 53.63 & 31.37 & 24.64 & 11.97 \\
        QD-DETR \cite{qddetr} & 57.31 & 32.55 & - & - \\
        UniVTG \cite{univtg} & 58.01 & 35.65 & 34.97 & 17.35 \\ 
        UVCOM \cite{uvcom} & 59.25 & 36.64 & 36.39 & 23.32 \\ 
        R2-tuning \cite{r2_tuning} & 59.78 & 37.02 & 38.72 & 25.12 \\ 
        \textbf{CoSTL (Ours)} & \textbf{62.04} &\textbf{40.13} & \textbf{39.21} & \textbf{25.69} \\ 
        \bottomrule
    \end{tabular*}
    \label{tab:MR}
    \vspace{-13pt}
\end{table}

% highlight实验: youtube(YoutubeHL) and tvsum
% \begin{table}[ht]
% \centering

% \begin{tabular}{lccccccc}
% \toprule
% \textbf{Method} & \textbf{Dog} & \textbf{Gym.} & \textbf{Par.} & \textbf{Ska.} & \textbf{Ski.} & \textbf{Sur.} & \textbf{Avg.} \\
% \midrule
% RRAE        & 49.0 & 35.0 & 50.0 & 25.0 & 22.0 & 49.0 & 38.3 \\
% GIFs        & 30.8 & 33.5 & 54.0 & 25.6 & 32.4 & 51.4 & 46.4 \\
% LSVM         & 37.9 & 41.0 & 67.0 & 37.4 & 36.7 & 65.0 & 48.3 \\
% LIM-S        & 57.9 & 41.1 & 67.0 & 57.7 & 57.7 & 70.2 & 58.6 \\
% SL-Module    & 70.8 & 53.2 & 77.2 & 72.5 & 66.1 & 76.7 & 69.3 \\
% \midrule
% MINI-Net$^\dagger$  & 58.2 & 61.7 & 70.2 & 70.2 & 58.7 & 65.1 & 64.4 \\
% TCG$^\dagger$      & 55.4 & 62.7 & 72.2 & 73.9 & 72.5 & 72.3 & 68.2 \\
% Joint-VAT            & 64.5 & 71.9 & 69.4 & 62.0 & 73.8 & 72.3 & 71.8 \\
% UMT$^\dagger$       & 65.9 & 75.2 & 81.6 & 71.7 & 74.1 & 82.7 & 74.9 \\
% \midrule
% UniVTG                   & 71.8 & 76.5 & 73.9 & 73.3 & 73.2 & 82.2 & 75.2 \\
% UniVTG w/ PT             & \textbf{74.3} & \textbf{79.0} & \textbf{74.4} & \textbf{74.9} & \textbf{75.1} & \textbf{83.9} & \textbf{76.9} \\
% UniVTG ZS                & 36.8 & 62.8 & 65.9 & 39.2 & 64.5 & 54.0 & 53.9 \\
% \midrule
% \textbf{Ours}                & ** & ** & ** & ** & ** & ** & ** \\
% \bottomrule
% \end{tabular}
% \caption{Performance comparison of YouTube HL across various types of videos}
% \label{tab:youtube_comparison}
% \end{table}

\begin{table}[h]
\centering
\caption{Comparisons on TVSum. † denotes using audio modality, the best and second-best metrics are marked with \textbf{bold} and \underline{underline}, respectively.}
% \resizebox{0.98\columnwidth}{!}{
% \begin{tabular}{l|cccccccccccc|c}
\begin{tabular*}{\columnwidth}{@{\extracolsep{\fill}}c|cccccccccccc|c}
\hline
\textbf{Method} & \textbf{VT} & \textbf{VU} & \textbf{GA} & \textbf{MS} & \textbf{PK} & \textbf{PR} & \textbf{FM} & \textbf{BK} & \textbf{BT} & \textbf{DS} & \textbf{Avg.} \\
\hline
sLSTM \cite{slstm} & 41.1 & 46.2 & 46.3 & 47.7 & 44.8 & 46.1 & 45.2 & 40.6 & 47.1 & 45.5 & 45.1 \\
LIM-S \cite{highlightless} & 55.9 & 42.9 & 61.2 & 54.0 & 60.4 & 47.5 & 43.2 & 66.3 & 69.1 & 62.6 & 56.3 \\
Trailer \cite{trailer}& 61.3 & 54.6 & 65.7 & 60.8 & 59.1 & 70.1 & 58.2 & 64.7 & 65.6 & 68.1 & 62.8 \\
SL-Module \cite{slmodule} & 86.5 & 68.7 & 74.9 & \underline{86.2} & 79.0 & 63.2 & 58.9 & 72.6 & 78.9 & 64.0 & 73.3 \\
MINI-Net† \cite{mini}& 80.6 & 68.3 & 78.2 & 81.8 & 78.1 & 65.8 & 57.8 & 75.0 & 80.2 & 65.5 & 73.2 \\
TCG† \cite{tcg}& 85.0 & 71.4 & 81.9 & 78.6 & 80.2 & 75.5 & 71.6 & 77.3 & 78.6 & 68.1 & 76.8 \\
Joint-VA† \cite{highlightjoint}& 83.7 & 57.3 & 78.5 & 86.1 & 80.1 & 69.2 & 70.0 & 73.0 & \textbf{97.4} & 67.5 & 76.3 \\
UniVTG \cite{univtg}& 83.9 & 85.1 & 89.0 & 80.1 & 84.6 & 81.4 & 70.9 & 91.7 & 73.5 & 69.3 & 81.0 \\
UMT† \cite{umt} & 87.5 & 81.5 & 88.2 & 78.8 & 81.5 & 87.0 & 76.0 & 86.9 & 84.4 & \textbf{79.6} & 83.1 \\
QD-DETR \cite{qddetr} & \textbf{88.2} & 87.4 & 85.6 & 85.0 & 85.8 & \underline{86.9} & 76.4 & \underline{91.3} & 89.2 & 73.7 & 85.0 \\

UVCOM \cite{uvcom} & \underline{87.6} & \underline{91.6} & \underline{91.4} & \textbf{86.7} & \underline{86.9} & 86.9 & 76.9 & \textbf{92.3} & 87.4 & 75.6 & \underline{86.3} \\
R2-tuning \cite{r2_tuning} & 85.0 &85.9& 91.0& 81.7& \textbf{88.8}& \textbf{87.4}& \underline{78.1}& 89.27 &\underline{90.3}& 74.7& 85.2  \\
\hline
\textbf{CoSTL (Ours)} & 86.6 & \textbf{91.7} & \textbf{92.6} & 84.9 & \underline{87.8} & 85.7 & \textbf{79.4} & 89.6 & 89.7 & \underline{77.3} & \textbf{86.5}\\
\hline
\end{tabular*}
\label{tab:hd}
\vspace{-13pt}
\end{table}

\textbf{Evaluation Metrics.} We adopt Recall@K, IoU=x,  mAP with IoU Thresholds as evaluation metrics. Specifically, Recall@K, IoU=x (R@K, IoU=x) evaluate the fraction of moments that are correctly retrieved with an Intersection over Union (IoU) greater than or equal to a threshold within the top K retrieved results. They assess the localization accuracy of temporal segments. mAP with IoU thresholds is similar to mAP in highlight detection but adapted for temporal moment retrieval, considering IoU thresholds to determine positive matches. For QVHighlights, we utilize Recall@1 with IoU thresholds 0.5 and 0.7, mean average precision (mAP) with IoU thresholds 0.5 and 0.75, and the average mAP over a series of IoU thresholds [0.5:0.05:0.95] are used for moment retrieval. For highlight detection, mAP and HIT@1 are used, a clip is treated as a true positive if it has a saliency score of Very Good. For TACoS, Recall@1 with IoU thresholds 0.3, 0.5 and 0.7, and mIoU are used. For TVSum, we use Top-5 mAP.

% \textbf{Implementation Details.}
% In all experiments, our image encoder uses Clip\_L \cite{clip} and Image Q-Former uses the pretrained weights in Blip2 \cite{blip2} for initialization. The temporal scales in multi-scale perception module are set as $\left \{ 1,2,4 \right \} $, the weight coefficients in the loss function are $\lambda _{f}=10$, $\lambda _{L_{1} }=10 $, $\lambda _{iou}=1$, $\lambda _{h}=1$. All experiments are implemented on a single A100-80g. Please see the Appendix for more training details.

\subsection{Comparison with State-of-the-arts}
% In this section, we compare our method with existing methods on four popular benchmarks across these two tasks.

\textbf{Joint Moment Retrieval and Highlight Detection.} 
We evaluate our method on QVHighlights dataset \cite{mdetr}, which supports video moment retrieval and highlight detection. We compare our approach, CoSTL, against state-of-the-art methods, including both pre-trained and non-pre-trained models. Because CoSTL initializes with BLIP-2 pre-trained weights, we also include a variant of the UVCOM \cite{uvcom} model using BLIP-2 features (denoted as UVCOM\textsuperscript{*}) for a fair comparison. As shown in Table \ref{tab:qvhighlight}, CoSTL achieves significant improvements across all metrics. Specifically, it surpasses the previous state-of-the-art, R2-tuning \cite{r2_tuning}, by 3.73\% on average across all metrics, demonstrating its effectiveness. 

\textbf{Moment Retrieval.} We evaluate our method on two benchmarks Charades-STA and TACoS. As shown in Table \ref{tab:MR}, our CoSTL outperforms previous method on both two benchmarks, demonstrating the superiority of the spatial-temporal features in CoSTL. Specifically on R1@0.5, our method increases 2.26\% in Charades-STA and increases 0.49\% in TACoS compared to R2-tuning.

\textbf{Highlight Detection.} We also evaluate our method on highlight detection benchmark TVSum in Table \ref{tab:hd}, although the number of samples in each category of TVSum is very small, our method still achieves good results in each category and achieves the best average performance 86.5\%.\footnote{Since we need raw video of datasets and YouTube HL \cite{youtbe} is proposed too early and too many videos are lost, we did not test metrics on it.}
%\subsection{Others}

\subsection{Ablation Studies}

\begin{table}[t]
    \centering
    \caption{Performance comparison for different modules. We ablated the importance of each module in MR and HD tasks.}
    % \renewcommand{\arraystretch}{1.0} % 增加行高
    % \resizebox{0.7\columnwidth}{!}{ % 自动调整表格宽度以适应列宽
    % \Huge
    % \begin{tabular}{cc|ccc|cc}
    \begin{tabular*}{\columnwidth}{@{\extracolsep{\fill}}cc|ccc|cc}
        % \toprule
        \hline
        \multirow{2}{*}{\textbf{TDE}} & \multirow{2}{*}{\textbf{MTP}} & \multicolumn{3}{c|}{\textbf{MR}} & \multicolumn{2}{c}{\textbf{HD}} \\
         &  & R1 @
         0.5 & R1 @0.7 & mAP Avg. & mAP & HIT@1 \\
        % \midrule
        \hline
        \checkmark &  & 68.81 & 43.38 & 39.89 & 40.18 & 70.73 \\
         & \checkmark & 72.05 & 47.88 & 42.05 & 38.75 & 65.17 \\
        \checkmark & \checkmark & \textbf{75.43} & \textbf{54.24} & \textbf{46.68} & \textbf{41.46} & \textbf{71.59} \\
        % \bottomrule
        \hline
    \end{tabular*}
    % }
    \label{tab:ablation1}
    \vspace{-13pt}
\end{table}
\begin{table}[b]
    \centering
    \vspace{-25pt}
    \caption{Ablation studies on text-driven progressive fine-grained encoder. }
    % \renewcommand{\arraystretch}{1.0}
    % \resizebox{0.7\columnwidth}{!}{
    % \Huge
    % \begin{tabular}{cc|ccc|cc}
    \begin{tabular*}{\columnwidth}{@{\extracolsep{\fill}}cc|ccc|cc}
        \hline
        \multirow{2}{*}{\textbf{Step1}} & \multirow{2}{*}{\textbf{Step2}} & \multicolumn{3}{c|}{\textbf{MR}} & \multicolumn{2}{c}{\textbf{HD}} \\
         &  & R1 @0.5 & R1 @0.7 & mAP Avg. & mAP & HIT@1 \\
        \hline
         & \checkmark & 70.39 & 43.38 & 39.38 & 38.75 & 65.17 \\
        \checkmark & M1 & 68.46 & 52.17 & 45.73 & 39.21 & 69.59 \\
        \checkmark & M2 & 70.81 & 52.38 & 45.89 & 40.18 & 70.73 \\
        \checkmark & \checkmark & \textbf{75.43} & \textbf{54.24} & \textbf{46.68} & \textbf{41.46} & \textbf{71.59} \\
        \hline
    \end{tabular*}
    % }
    \label{tab:ablation2}
\end{table}
% . The results demonstrate that step 1 and step 2 are indispensable in our method.
% In this section, we conduct a comprehensive analysis of our proposed modules TDE and MTP.

 \textbf{Module analysis.} 
% Specifically, to valiadate the effectiveness of MCL, we discard the MTP module and directly feed the features $F_{v}^{''}$ into the prediction heads. 
% As for TDE, in order to verify that the improvement in performance comes from the design of the TCE module rather than the enhancement of features, we change this module to directly use the BLIP2 feature followed by MTP. 
We analyze the effectiveness of our proposed Text-driven Progressive Fine-grained Image Encoder (TDE) and Multi-scale Temporal Perception Module (MTP). To evaluate the contribution of the MTP module, we conduct an ablation study where we remove the MTP module and directly feed the features $F_v^{''}$ into the prediction heads. 
As for TDE, to confirm that the performance improvement stems from the design of the TDE module and not simply from enhanced visual features, we replace the TDE module with the standard BLIP-2 features, which are then processed by the MTP module. Table \ref{tab:ablation1} presents the ablation results. Both the MTP and TDE modules individually contribute to improved performance, and their combination yields even better results, such as a 7.30\% increase in mAP(MR). More specifically, the MTP module demonstrates a significant impact on moment retrieval metrics, exemplified by a 6.62\% improvement in R1@0.5, validating the effectiveness of its multi-scale temporal modeling. Conversely, the TDE module shows a greater impact on highlight detection, with a 6.42\% increase in HIT@1, demonstrating its ability to extract fine-grained spatial information.

\textbf{Text-driven progressive fine-grained encoder} 
TDE progressively refines spatial feature embeddings to be more precise and relevant to the text query through a two-step interaction process. To evaluate the effectiveness of each step, we conduct two experiments. Firstly, to assess the effectiveness of valid spatial information extraction in Step 1, we replaced this step with standard BLIP-2 features. Then to evaluate the effectiveness of frame information aggregation in Step 2, we compared several alternative aggregation methods, including standard pooling (M1) and an attention mechanism without text-based enhancement (M2). From Table \ref{tab:ablation2}, we can see that both Step 1 and Step 2 play an important role on performance improvement, e.g.,  5.04\% increase in R1@0.5 for Step 1. The comparison results with other aggregation strategies also show the effectiveness of our Step 2 design, e.g., 6.98\% increase in R1@0.5 vs M1 and 4.62\% increase in R1@0.5 vs M2.

\begin{figure*}[t]
    \centering
    \begin{subfigure}{\linewidth}
        \centering
        \includegraphics[width=\linewidth]{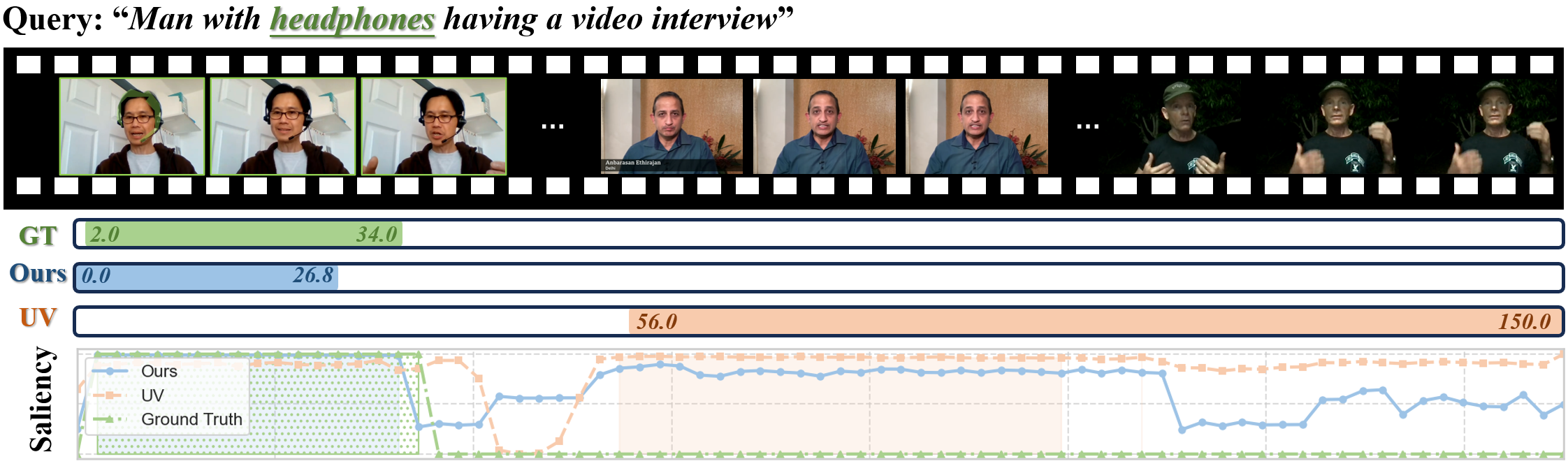}
        \label{fig:vis_spa}
    \end{subfigure}
    % \vskip\baselineskip
    \begin{subfigure}{\linewidth}
        \centering
        \includegraphics[width=\linewidth]{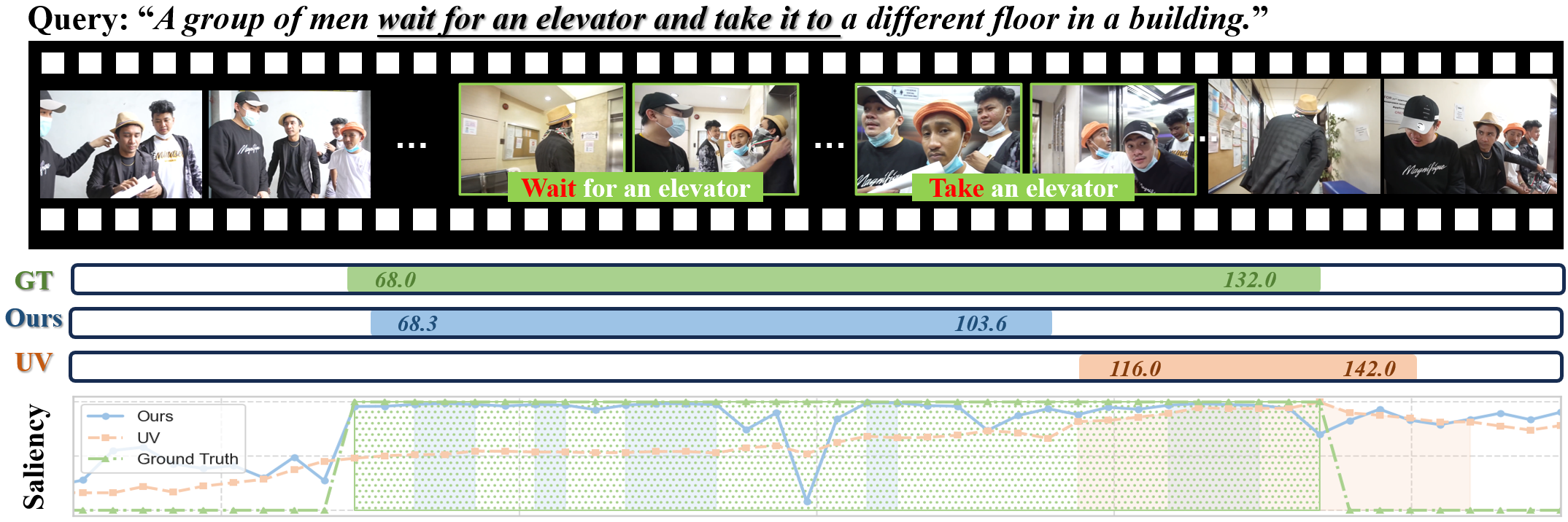}
        \label{fig:vis_temp}
    \end{subfigure}
    \caption{More visualizations of joint moment retrieval and highlight detection results on QVHighlights val split. Our method can accurately regress the boundaries of moments and predict highlight saliency scores through its novel design.}
    \label{fig:combined}
    \vspace{-13pt}
\end{figure*}

\textbf{Multi-scale temporal perception module} MTP captures temporal dynamics in videos through multi-scale temporal aggregation, thereby obtaining spatial-temporal features for subsequent prediction. 
To investigate the role of multi-scale temporal perception, we conducted experiments with different scale configurations: $S = {1}$ (single-scale) and $S = {1, 2}$ (dual-scale). The single-scale setting serves as a baseline without multi-scale information, while the dual-scale setting aggregates information at two temporal scales. As Table \ref{tab:ablation1} demonstrates, the MTP module primarily affects moment retrieval performance. Therefore, we focus our analysis on moment retrieval metrics. Table \ref{tab:ablation3} shows that our multi-scale design leads to improved performance. Specifically, using multi-scale aggregation ($S = {1, 2, 3}$) results in a 3.97\% increase in R1@0.5 compared to the single-scale baseline ($S = {1}$). Furthermore, our full multi-scale approach outperforms dual-scale aggregation ($S = {1, 2}$) by 1.39\% in R1@0.5.

% ablation study中的表格
% \begin{table}[h]
%     \centering
%     \caption{Analysis of different temporal scale $S$ in MTP, where TS denotes number of temporal scales}
%     \resizebox{\columnwidth}{!}{
%     \begin{tabular}{ccccc}
%         \toprule
%         number & \multicolumn{4}{c}{MR}  \\ \cmidrule(lr){2-5} 
%         of TS &R1@0.3 &R1@0.5 & R1@0.7 & mAP   \\ \midrule
%         1 &85.43 & 71.46 & 53.17 & 45.43  \\
%         2 & 87.75 & 74.04 & 53.83 & 45.73  \\ 
%         3 & 88.28 & 75.43 & 54.24 & 46.68   \\ 
%         \bottomrule
%     \end{tabular}}
%     \label{tab:ablation3}
% \end{table}

\begin{table}[b]
    \centering
    \vspace{-13pt}
    \caption{Analysis of different temporal scale $S$ in MTP, where TS denotes number of temporal scales}
    \begin{tabular*}{\columnwidth}{@{\extracolsep{\fill}}ccccc}
        \toprule
        \multicolumn{1}{c}{\textbf{Number}} & \multicolumn{4}{c}{\textbf{MR}}  \\ \cmidrule(lr){2-5} 
        \multicolumn{1}{c}{\textbf{of TS}} & R1@0.3 & R1@0.5 & R1@0.7 & mAP   \\ \midrule
        1 &85.43 & 71.46 & 53.17 & 45.43  \\
        2 & 87.75 & 74.04 & 53.83 & 45.73  \\ 
        3 & 88.28 & 75.43 & 54.24 & 46.68   \\ 
        \bottomrule
    \end{tabular*}
    \label{tab:ablation3}
\end{table}
\subsection{Qualitative Results}
%We show more visualizations of qualitative results in Fig. \ref{fig:combined}. Compare with the strong baseline UVCOM \cite{uvcom}. Our method can regress moment boundaries and detect highlights more accurately due to the novel design. In the first case, we show an example when facing complex queries with detailed descriptions, such as people's outfit, colors, etc. Our method is able to more accurately locate the details in the text into the frames. In the second case, Our approach is not only able to separate the details from the picture, but also to understand the action content across a larger time dimension.

We show more visualizations of qualitative results in Fig. \ref{fig:combined}, comparing our method with the strong baseline UVCOM \cite{uvcom}. Our method demonstrates improved accuracy in both moment boundary regression and highlight detection.

In the first example, showcasing a complex query with detailed descriptions of people's attributes ``headphones'', our method precisely locates the corresponding visual details within the video frames, outperforming UVCOM which struggles with such fine-grained queries. 

The second example highlights our method's ability to not only extract detailed visual information but also understand the temporal dynamics of actions across a broader time span, which can capture ``Wait for an elevator'' and ``Take an elevator'' simultaneously.l approach might falter. 
UVCOM, in contrast, misses the overall context of the action. These examples demonstrate the effectiveness of our proposed spatial-temporal modeling approach in handling both fine-grained details and broader temporal context.
\section{Conclusion}
This paper introduces CoSTL, a comprehensive spatial-temporal representation learning framework for video moment retrieval and highlight detection. Existing methods often struggle to capture fine-grained spatial information and effectively model temporal dynamics, leading to inaccurate results. CoSTL addresses these limitations by tackling two key challenges: (1) Capturing fine-grained visual details related to the text within individual frames. (2) Effectively modeling the dynamic temporal relationships within a video. We introduced specialized modules designed to address each of these challenges. Experimental results demonstrate the effectiveness of our proposed method.

\bibstyle{splncs04}
\bibliography{main}
\end{document}